\PassOptionsToPackage{table}{xcolor}
\documentclass[lettersize,journal]{IEEEtran}
\usepackage{tikz}
\usepackage{pgfplots}
\usepackage{subcaption}
\usepackage{stmaryrd}
\usepackage{amsmath,amsfonts,amssymb, bm}
\usepackage{algorithm}
\usepackage{algorithmicx}
\usepackage[noend]{algpseudocode}
\usepackage{enumitem}
\usepackage{array}
\usepackage[caption=false,font=normalsize,labelfont=sf,textfont=sf]{subfig}
\usepackage[table]{xcolor}
\usepackage{textcomp}
\usepackage{stfloats}
\usepackage{url}
\usepackage{verbatim}
\usepackage{graphicx}
\usepackage{cite}
\usepackage{xcolor}
\usepackage{booktabs}
\usepackage{multirow}
\pgfplotsset{compat=1.18}
\usepackage{booktabs}
\usepackage{tabularx}
\usepackage{array}
\newcolumntype{Y}{>{\raggedright\arraybackslash}X} 

\definecolor{dataset1}{RGB}{228, 26, 28}
\definecolor{dataset2}{RGB}{55, 126, 184}
\definecolor{dataset3}{RGB}{77, 175, 74}
\definecolor{color1}{RGB}{69, 123, 157}     
\definecolor{color2}{RGB}{229, 89, 52}      
\definecolor{color3}{RGB}{144, 190, 109}    
\definecolor{color4}{RGB}{255, 159, 28}     
\definecolor{fontgreen}{RGB}{0, 128, 2}  

\begin{document}

\title{Routing Distilled Knowledge via Mixture of LoRA Experts for Large Language Model based Bundle Generation}

\author{Kaidong Feng, Zhu Sun, Hui Fang, Jie Yang, Wenyuan Liu, \\Yew-Soon Ong \textit{IEEE Fellow}
\IEEEcompsocitemizethanks{
\IEEEcompsocthanksitem Zhu Sun (corresponding author) is with Singapore University of Technology and Design, Singapore. E-mail:sunzhuntu@gmail.com
\IEEEcompsocthanksitem Kaidong Feng and Wenyuan Liu are with Yanshan University, China. E-mail:fengkaidong@stumail.ysu.edu.cn; wyliu@ysu.edu.cn 
\IEEEcompsocthanksitem Jie Yang is with Delft University of Technology, the Netherlands. E-mail: j.yang-3@tudelft.nl
\IEEEcompsocthanksitem Hui Fang is with Shanghai University of Finance and Economics, China. E-mail: fang.hui@mail.shufe.edu.cn
\IEEEcompsocthanksitem Yew-Soon Ong is with A*STAR Centre for Frontier AI Research, Singapore. E-mail: ong\_yew\_soon@hq.a-star.edu.sg. He is also with Nanyang Technological University, Singapore. E-mail: asysong@ntu.edu.sg
}
\thanks{© 2025 IEEE. Personal use of this material is permitted. 
    Permission from IEEE must be obtained for all other uses, 
    in any current or future media, including reprinting/republishing 
    this material for advertising or promotional purposes, 
    creating new collective works, for resale or redistribution 
    to servers or lists, or reuse of any copyrighted component 
    of this work in other works.}
}

\markboth{Journal of \LaTeX\ Class Files,~Vol.~14, No.~8, August~2021}%
{Shell \MakeLowercase{\textit{et al.}}: A Sample Article Using IEEEtran.cls for IEEE Journals}


\maketitle

\begin{abstract}
Large Language Models (LLMs) have shown potential in automatic bundle generation but suffer from prohibitive computational costs. Although knowledge distillation offers a pathway to more efficient student models, our preliminary study reveals that naively integrating diverse types of distilled knowledge from teacher LLMs into student LLMs leads to knowledge conflict, negatively impacting the performance of bundle generation.
To address this, we propose RouteDK, a framework for routing distilled knowledge through a mixture of LoRA expert architecture. Specifically, we first distill knowledge from the teacher LLM for bundle generation in two complementary types: high-level knowledge (generalizable rules) and fine-grained knowledge (session-specific reasoning). We then train knowledge-specific LoRA experts for each type of knowledge together with a base LoRA expert. For effective integration, we propose a dynamic fusion module, featuring an input-aware router, where the router balances expert contributions by dynamically determining optimal weights based on input, thereby effectively mitigating knowledge conflicts.
To further improve inference reliability, we design an inference-time enhancement module to reduce variance and mitigate suboptimal reasoning.
Experiments on three public datasets show that our RouteDK achieves accuracy comparable to or even better than the teacher LLM, while maintaining strong computational efficiency. In addition, it outperforms state-of-the-art approaches for bundle generation. 
\end{abstract}


\section{Introduction}\label{sec:intro}

\IEEEPARstart{P}{roduct} bundling is a critical merchandising strategy that groups a number of  complementary or alternative items into a single package, applied in various domains such as e-commerce, retail, and telecommunications~\cite{sar2016beyond,dragone2018no,sun2022revisiting}. With this strategy, vendors can satisfy diverse customer needs, enhance user experiences, and drive increased sales and engagement, while users benefit from convenience and discounted price. Therefore, more and more researchers pay attention to this topic, especially in a task known as \textbf{bundle generation}~\cite{sun2022revisiting, wei2022towards,sun2024adaptive}. It aims to automatically construct appealing bundles from a set of products (i.e., user session) by inferring user intents. These generated bundles directly benefit the downstream tasks, such as bundle recommendation~\cite{bai2019personalized,chang2021bundle,he2022bundle}, by supplying better initial candidate bundles.

Traditionally, bundle generation has been approached through constraint-based methods or neural network models, which rely on maximizing objectives like cost saving and expected revenue, or modeling the implicit relationships between items and assemble bundles~\cite{xie2010breaking,beladev2016recommender,zhu2014bundle}. However, these conventional methods face several critical challenges. For instance, they often struggle to guarantee diversity and support flexible bundle sizes due to inherent limitations in modeling flexibility and expressiveness. 

Recent advances in large language models (LLMs) present promising solutions to these challenges. LLMs possess exceptional capabilities in semantic understanding, reasoning, and generating contextually coherent content~\cite{wang2023generative, wang2023enhancing}. Such strengths enable them to better capture product relationships, and infer user intent, thus addressing the limitations of traditional bundle generation methods. 
Nevertheless, adopting LLMs for bundle generation presents notable challenges, primarily concerning computational efficiency. High-performing LLMs, while effective, typically require significant computational resources, limiting their practical scalability and deployment in real-world systems. To address this challenge, recent research~\cite{feng2025does} has explored knowledge distillation techniques. Specifically, it explicitly distills knowledge from large, more capable teacher LLMs to fine-tune smaller, more efficient student LLMs, thereby achieving a balance between effectiveness and computational efficiency.

\begin{figure}
\centering
\begin{subfigure}{0.48\columnwidth}
    \begin{tikzpicture}
        \begin{axis}[
            width=5cm,
            height=4.5cm,
            ylabel={Precision},
            ylabel style={
                color=black,
                at={(axis description cs:0.5,1.05)},
                anchor=south,  
                rotate=270
            },
            ymin=0.595, ymax=0.660,
            xtick={1,2,3,4},
            xticklabels={$\text{DK}_1$, $\text{DK}_2$, $\text{DK}_3$, $\text{DK}_4$},
            legend style={at={(0.33,1.00)}, font=\footnotesize,         anchor=north,legend columns=1, draw=none, fill=none,
            },
            grid=none,
            grid style={dashed,gray!30},
            tick label style={font=\footnotesize},
            label style={font=\small},
        ]
        
        \addplot[
            color=color1,
            mark=square*,
            mark size=1.5pt,
            line width=1pt,
        ] coordinates {
            (1, 0.618) (2, 0.610) (3, 0.633) (4, 0.619) 
        };
        \addlegendentry{Electronic}
        
        \addplot[
            color=color2,
            mark=triangle*,
            mark size=1.5pt,
            line width=1pt,
        ] coordinates {
            (1, 0.608) (2, 0.609) (3, 0.614) (4, 0.619) 
        };
        \addlegendentry{Clothing}
        
        \addplot[
            color=color3,
            mark=diamond*,
            mark size=1.5pt,
            line width=1pt,
        ] coordinates {
            (1, 0.615) (2, 0.623) (3, 0.635) (4, 0.648) 
        };
        \addlegendentry{Food}
        
        \end{axis}
    \end{tikzpicture}
\end{subfigure}
\hfill
\begin{subfigure}{0.48\columnwidth}
    \begin{tikzpicture}
\begin{axis}[
    width=5cm,
    height=4.5cm,
    ylabel={Recall},
    ylabel style={
    color=black,
    at={(axis description cs:0.5,1.05)},
    anchor=south,  
    rotate=270
    },
    ymin=0.585, ymax=0.660,
    xtick={1,2,3,4},
    xticklabels={$\text{DK}_1$, $\text{DK}_2$, $\text{DK}_3$, $\text{DK}_4$},
    legend style={at={(0.3, 1.00)}, font=\footnotesize,         anchor=north,legend columns=1, draw=none, fill=none,
    },
    yticklabel pos=right,
    grid=none,
    grid style={dashed,gray!30},
    tick label style={font=\footnotesize},
    label style={font=\small},
]

\addplot[
    color=color1,
    mark=square*,
    mark size=1.5pt,
    line width=1pt,
] coordinates {
    (1, 0.621) (2, 0.594) (3, 0.644) (4, 0.640) 
};
\addlegendentry{Electronic}

\addplot[
    color=color2,
    mark=triangle*,
    mark size=1.5pt,
    line width=1pt,
] coordinates {
    (1, 0.607) (2, 0.598) (3, 0.621) (4, 0.620) 
};
\addlegendentry{Clothing}

\addplot[
    color=color3,
    mark=diamond*,
    mark size=1.5pt,
    line width=1pt,
] coordinates {
    (1, 0.597) (2, 0.600) (3, 0.640) (4, 0.639) 
};
\addlegendentry{Food}
\end{axis}
\end{tikzpicture}
\end{subfigure}
\caption{Impact of distilled knowledge (DK) types during fine-tuning on Precision and Recall across the Electronic, Clothing, and Food domains. $\text{DK}_1$: \textit{Raw data}, without knowledge injection; $\text{DK}_2$: \textit{Rule}, with injection of high-level rules; $\text{DK}_3$: \textit{Thought}, with injection of fine-grained reasoning; $\text{DK}_4$: \textit{Merged}, with both rules and reasoning injected.}
\label{fig:knowledge_comb}
\vspace{-0.15in}
\end{figure}
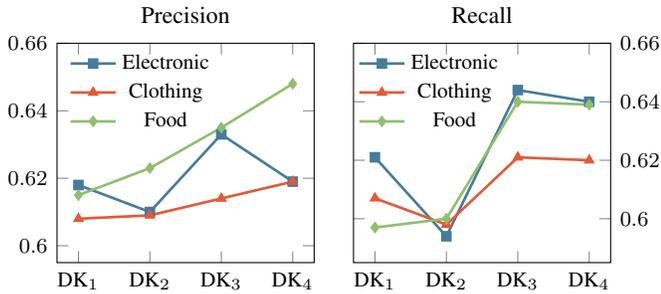

Despite the achievements, it naively integrates different types of distilled knowledge from teacher LLMs, which may lead to knowledge conflicts and consequently degrade performance. To investigate this, we conducted a preliminary study by fine-tuning LLMs where distilled knowledge of different types were naively concatenated as in~\cite{feng2025does}. The results, presented in Figure~\ref{fig:knowledge_comb}, reveal a consistent trend across three datasets: this simple aggregation strategy not only fails to improve performance but often degrades it below the level achieved by using a single knowledge type. This finding suggests the presence of \textit{knowledge conflict}, where the undifferentiated mixture of knowledge types hinders—rather than facilitates—the model’s reasoning process. It highlights the need for a more effective approach to knowledge fusion.


To this end,  we propose RouteDK, a novel framework that dynamically routes distilled knowledge through a mixture of LoRA expert architecture to further enhance the effectiveness of a lightweight LLM on the bundle generation task. Specifically, we first distill two types of complementary knowledge from the teacher LLM: \textit{high-level} knowledge (generalizable rules), and \textit{fine-grained} knowledge (session-specific reasoning) for bundle generation. Then, to better accommodate different types of distilled knowledge, we design three specialized LoRA-based experts: a \textit{Base Expert} for foundational capabilities, a \textit{High-level Knowledge Expert} for generalizable rules, and a \textit{Fine-grained Knowledge Expert} for session-specific reasoning. To effectively integrate them, we propose a novel \textit{Dynamic Fusion} module that exploits a lightweight router to dynamically allocate weights for different types of knowledge based on the input session. 
By explicitly separating knowledge acquisition (in the LoRA experts) from its utilization (via the dynamic fusion), our model mitigates knowledge conflicts and synthesizes information more effectively. Lastly, we further improve the reliability of inference  through a test-time scaling strategy based on self-consistency. Extensive experiments on three public datasets confirm that our framework achieves substantial performance gains while maintaining high computational efficiency.

Overall, our contributions can be summarized as: 
\begin{itemize}
    \item We identify the knowledge conflict issue in existing  knowledge distillation based bundle generation methods, where naively merging different types of distilled knowledge as context for lightweight LLMs leads to performance degradation.
    \item  We propose RouteDK, a novel framework designed to resolve such knowledge conflict issue for more effective and efficient bundle generation. Specifically, it first distills two types of complementary knowledge from the teacher LLMs, which are then accommodated by the specialized LoRA experts for learning. Subsequently, a novel dynamic fusion module is design to effectively integrate their impacts. Finally, during inference, it further incorporates a test-time scaling strategy to improve the robustness of the final results.
    \item Experiments on three public datasets show that RouteDK achieves accuracy comparable to, and in some cases exceeding, that of the teacher LLM, while maintaining high computational efficiency. Moreover, it outperforms state-of-the-art bundle generation methods, with an average gain of 9.32\% over knowledge distillation based ones. 
\end{itemize}

\section{Related Works}
\subsection{Bundle Generation}
Bundle generation aims to compose sets of complementary or alternative items that satisfy a user’s intent~\cite{sun2022revisiting}, where research has progressed from early co-occurrence mining to personalized and multimodal modeling. Conventional-based methods first mine frequent patterns~\cite{sun2022revisiting} at the category level for bundle generation. It then shifts to personalized bundle generation by learning compatibility together with user preferences like BYOB~\cite{deng2021build}, which employs reinforcement learning to sequentially build personalized bundles, guided by a reward function that balances user preference with item compatibility. The recent work exploits multimodal inputs (i.e., image and text) and user historical interactions.  POG~\cite{chen2019pog} generates personalized outfits using a Transformer architecture that conditions its output on multi-modal item embeddings derived from visual, textual, and collaborative signals; CLHE~\cite{ma2024leveraging} fuses heterogeneous modalities via self-attention; Conna~\cite{wei2022towards} introduces a contrastive non-autoregressive decoder to raise both creativity and efficiency; CIRP~\cite{ma2024cirp} injects cross-item semantics and relations into multimodal representations; and DiFashion~\cite{xu2024diffusion} uses diffusion to synthesize visually compatible fashion items. Despite their progress, these approaches usually demand substantial amounts of training data to learn stable item relations and user preferences, and they often produce fixed-size bundles or rely on partial bundles as input, which reduces their flexibility across scenarios.

These limits motivate moving beyond purely data-driven modeling toward methods that can reuse stronger prior knowledge and generalize with fewer labels. LLM-based methods follow this direction. Since LLMs carry broad prior knowledge, they can learn bundle–item relations from only a small amount of high-quality data when paired with light alignment. In particular, BundleLLM~\cite{liu2025harnessing} adds a trainable fusion module that aligns visual and textual features in the LLM semantic space, enabling mixed-modality processing and completion of partial bundles. AICL~\cite{sun2024adaptive} applies in-context learning to infer user intents and generate bundles with dynamic lengths, directly relaxing the fixed-size and partial-context constraints. However, LLM-based methods inevitably incur substantial computational overhead and response latency, limiting their practicality in real-world deployment.

\subsection{Knowledge Distillation in LLM-based Recommendation}
LLM-based recommendation leverages the powerful capabilities of LLMs by formulating recommendation as a natural language processing task~\cite{wu2023survey, sun2024large}. Existing methods follow two main paradigms: \textit{prompting-based methods} and \textit{tuning-based methods}. The former designs detailed instructions to guide LLMs in understanding recommendation tasks~\cite{zhai2023knowledge,dai2023uncovering,he2023large}, while the latter constructs training data to fine-tune LLMs for task-specific adaptation~\cite{bao2023tallrec,lin2024rella}. Although these methods have shown promising results, their efficiency is often limited by the substantial computational cost of LLMs.

Therefore, knowledge distillation (KD) has been introduced to transfer knowledge from large, complex teacher LLMs to lightweight, efficient student models. Current KD-based approaches can be divided into two categories. \textit{Explicit knowledge distillation} uses the outputs of teacher models as supervisory signals to guide student models toward generating similar content. For example, SLIM~\cite{wang2024SLIM} and RDRec~\cite{wang2024rdrec} leverage teacher-generated rationales behind user behaviors to enhance the reasoning ability of student models. In contrast, \textit{implicit knowledge distillation} improves student feature representations by aligning teacher and student outputs at hidden or output layers. For example, LaMP~\cite{salemi2024lamp} and PRM-KD~\cite{wen2024prmkd} minimize the KL-divergence between the probability distributions of the student and teacher for personalized text generation and item ranking. Other models further enhance effectiveness by jointly aligning intermediate representations and output logits, such as NewsBERT~\cite{wu2021newsbert}, Tiny-Newsrec~\cite{yu2021tiny}, and SSI~\cite{yuan2021improving}. 

For bundle generation task, KD4BG~\cite{feng2025does} distills different forms of knowledge from a large teacher LLM to a compact student LLM. However, its simple aggregation of all distilled knowledge leads to internal conflicts, leaving a notable effectiveness gap compared to the teacher. This challenge motivates our work, which focuses on resolving knowledge conflicts to build a student model that is both efficient and highly effective.

\subsection{Mixture of LoRA Experts}
The Mixture-of-Experts (MoE) framework~\cite{zhou2022moe} enhances efficiency by activating only a subset of experts per input, while Low-Rank Adaptation (LoRA)~\cite{dettmers2023qlora} enables parameter-efficient fine-tuning (PEFT)~\cite{fu2023peft} through lightweight low-rank updates. Recent research combines these ideas by constructing mixture of LoRAs~\cite{wu2024mixture}, where multiple LoRA adapters are treated as experts. Two main paradigms have emerged.
The first is \textit{static parameter merging}, which consolidates adapters into a single parameter set before inference to maximize efficiency, exemplified by AdaMix~\cite{wangetal2022adamix} and SMEAR~\cite{muqeeth2023soft}. The second is \textit{dynamic output ensembling}, which utilizes a router to activate and aggregate results from multiple LoRA specialists during inference (e.g., MoLoRA~\cite{zadouripushing}). Inspired by this, our work designs a mixture of LoRA experts where each adapter specializes in a distinct type of distilled knowledge, and their outputs are dynamically fused to mitigate knowledge conflicts in bundle generation.

\subsection{Test-time Scaling}
Test-time scaling refers to the use of additional computation to improve the performance of LLMs during inference. Existing studies involve three main branches. \textit{Parallel-based methods} generate several answers independently and select the optimal one via majority voting or an external reward model~\cite{brown2024large}. \textit{Sequential-based methods} iteratively refine solutions, where each step leverages earlier outputs to enable progressive error correction and improvement~\cite{snell2024scaling,hout1}. \textit{Hybrid-based methods} integrate parallel exploration and sequential refinement within a search tree. This family spans techniques from Monte-Carlo Tree Search~\cite{liudon} to guided beam search and advanced frameworks like REBASE~\cite{wu2024inference}, which has been shown to surpass traditional baselines using process-level reward signals. In this work, we follow parallel-based methods, where we explore the majority voting approach inspired by the principle of self-consistency. This strategy reduces the variance of single-shot generation and mitigates unreliable reasoning paths, while its simplicity ensures that the overall efficiency of the model is preserved.

\section{Preliminary}
\subsection{LLM-based Bundle Generation}
Given a session $s$ with a set of items $\mathcal{V}_s=\{v_1,v_2,\ldots,v_{|s|}\}$, each item $v_j$ is described by textual attributes (e.g., title, category). A \emph{bundle} is defined as a subset of at least two items. The task is to build a model that takes all items $\mathcal{V}_s$ in a session $s$ as input, and generates $M$ bundles: $\mathcal{B}_s=\{\,b_1,\dots,b_M\,\},\text{where}\ b_m \subseteq \mathcal{V}_s,\; |b_m|\ge 2 \ (m=1,\dots,M)$.
To adapt this task for LLMs, we formulate it as a text-to-text problem. The input session is serialized into a textual prompt, and the LLM is trained to generate an output sequence that textually represents the target bundles. This sequence-to-sequence formulation allows us to directly fine-tune the LLM on the task using its native text-generation capabilities. Important notations are summarized in Table ~\ref{tab:notation}.

\begin{table}[t]
  \centering
  \footnotesize
  \setlength{\tabcolsep}{4pt}
  \renewcommand{\arraystretch}{1.2}
  \caption{Important notations}
  \vspace{-0.1in}
  \label{tab:notation}
  \begin{tabularx}{\linewidth}{@{} l Y @{}}
    \toprule
    \textbf{Notation} & \textbf{Description} \\
    \midrule
    $\mathcal{S}$, $s\in\mathcal{S}$ & Set of training sessions; session $s$. \\
    $\mathcal{V}_s\text{=}\{v_1,\ldots,v_{|s|}\}$ & Item set in session $s$. \\
    $b_m$ & Bundle $m$ where $b_m \subseteq \mathcal{V}_s$ and $\lvert b_m\rvert \ge 2$. \\
    $\mathcal{B}_s\text{=}\{b_1,\ldots,b_M\}$ & Target bundle set for session $s$. \\
    $\mathcal{E}\text{=}\{e_{\mathrm{B}}, e_{\mathrm{H}}, e_{\mathrm{F}}\}$ & Set of LoRA experts, including base, high-level and fine-grained experts.\\
    $\boldsymbol{\theta}$ & Parameters of the frozen backbone LLM. \\
    $\boldsymbol{\phi_{e_i}}$ & Trainable parameters of LoRA expert $e_i$. \\
    $\boldsymbol{\psi}$ & Router network parameters for dynamic fusion. \\
    $x^s$ & {Raw input for session $s$ without distilled knowledge.} \\
    $x^{s}_{e_i}$ & Expert-specific input for session $s$. \\
    $y^s$ & Ground-truth textual sequence for session $s$. \\
    $y^s_{\tau}$ & Token at position $\tau$ in $y^s$. \\
    $y^s_{<\tau}$ & Prefix of $y^s$ before position $\tau$. \\
    $\mathbf{z}^{l}$ & Context vector for routing at layer $l$. \\
    $\boldsymbol{\alpha}^{l}_{e_i}\!\left(\mathbf{z}^{l}\right)$ & Router-assigned weight for expert $e_i$ at layer $l$. \\
    $\Delta^{l}_{e_i}\!\left(\mathbf{H}^{l-1}_{\mathrm{dyn}}\right)$ & LoRA update contributed by expert $e_i$ at layer $l$. \\
    $\mathbf{X}^s$ & Representation of tokenized $x^s$. \\
    $\mathbf{H}^{l}_{\mathrm{dyn}}$ & Hidden state at layer $l$ after dynamic fusion. \\
    \bottomrule
  \end{tabularx}
\end{table}

\subsection{LoRA-based Fine-tuning}
LoRA~\cite{dettmers2023qlora} is a parameter-efficient fine-tuning technique for LLMs. The core idea is to freeze the original model weights and inject trainable, low-rank matrices into specific layers of the LLM. This approach allows the LLM to adapt to domain-specific tasks with significant performance improvements by only training a very small number of new parameters.
Formally, let $\mathbf{H}^0=\mathbf{X} \in \mathbb{R}^{T \times d_{\text{in}}}$ denote the initial input sequence with length $T$ and hidden dimension $d_{\text{in}}$. For a target linear layer $l$ with its original weight $\mathbf{W}_{\boldsymbol{\theta}}^l \in \mathbb{R}^{d_{\text{in}} \times d_{\text{out}}}$, the forward pass takes the output from the previous layer, $\mathbf{H}^{l-1}$, as its input. The computation can be modified as:
\begin{equation}
    \mathbf{H}^l = f_{\boldsymbol{\theta}}^l(\mathbf{H}^{l-1}) + \Delta(\mathbf{H}^{l-1}; \boldsymbol{\phi}^{l}),
\end{equation}
where $f_{\boldsymbol{\theta}}^l(\mathbf{H}^{l-1}) = \mathbf{H}^{l-1}\mathbf{W}_{\boldsymbol{\theta}}^l$ represents the output from the original, frozen part of layer $l$. The term $\Delta(\mathbf{H}^{l-1}; \boldsymbol{\phi}^{l}) = \mathbf{H}^{l-1}\mathbf{A}^l \mathbf{B}^l$ represents the update from the LoRA module at layer $l$, where $\mathbf{A}^l\in\mathbb{R}^{d_{\text{in}}\times r}$ and $\mathbf{B}^l\in\mathbb{R}^{r\times d_{\text{out}}}$ are two trainable low-rank matrices. The rank $r$ is a hyperparameter and $r\ll \min\!\big(d_{\text{in}},\, d_{\text{out}}\big)$, which ensures the number of trainable parameters in $\mathbf{A}^l$ and $\mathbf{B}^l$ is minimal.
 
\section{Methodology}

\begin{figure*}[t]
    \centering
    \includegraphics[width=0.95\linewidth]{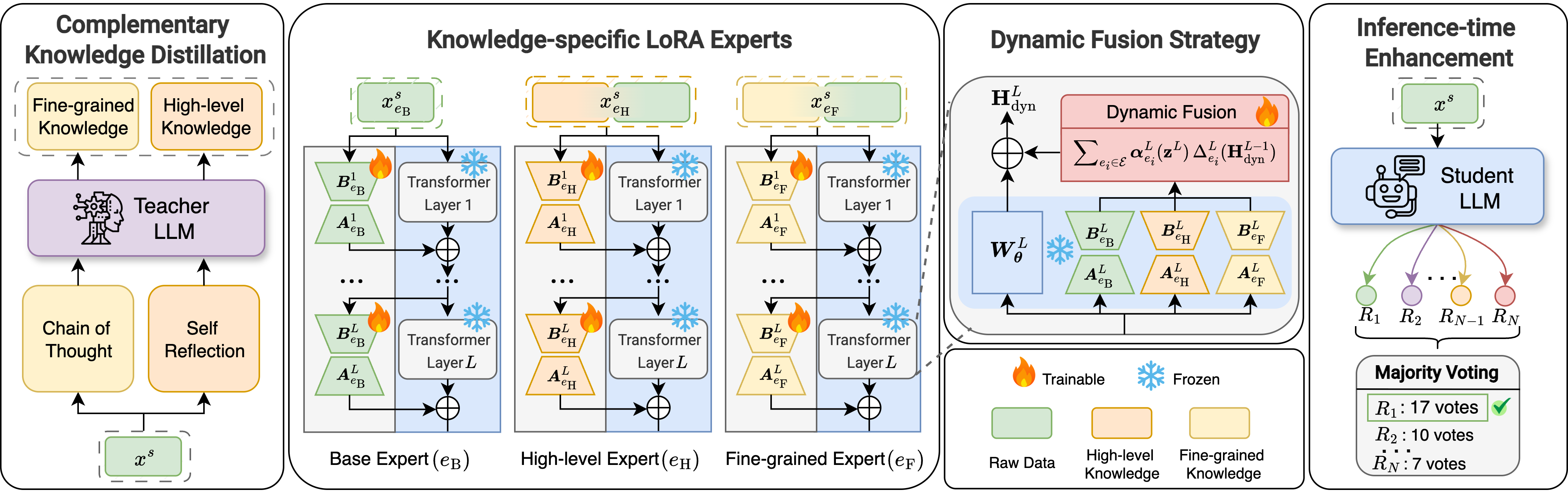}
    \caption{Illustration of the proposed RouteDK framework. }
    \label{fig:framework}
    \vspace{-0.15in}
\end{figure*}

In this section, we present our framework, RouteDK, which consists of four key components as shown in Figure~\ref{fig:framework}. First, we distill two complementary types of knowledge (i.e., high-level generalizable rules and fine-grained session-specific reasoning) from a teacher LLM for bundle generation. Second, we design knowledge-specific LoRA experts and train them by accommodating different types of distilled knowledge. Third, we introduce a dynamic fusion strategy to effectively integrate the outputs of these experts for more effective bundle generation. Finally, we propose an inference enhancement method that further improves the inference robustness through test-time scaling.

\subsection{Complementary Knowledge Distillation}
\label{sec:distill}

Directly fine-tuning lightweight LLMs on bundle generation data often fails to achieve satisfactory performance~\cite{feng2025does}. The primary reason is that small-scale LLMs lack the capacity to internalize both generalizable patterns across sessions and the nuanced reasoning required within individual sessions. This leads to unstable predictions and limited transferability. To address this issue, we employ knowledge distillation (KD) to transfer knowledge from the powerful teacher LLM to the student LLM for more efficient yet effective bundle generation.

Inspired by the prior work~\cite{feng2025does}, we distill two complementary types of knowledge that capture different aspects of the bundle generation process:

\begin{itemize}
    \item \textbf{High-level Knowledge.} This type of knowledge reflects generalizable rules governing bundle formation. To fully leverage the capabilities of the teacher LLM, we employ a self-reflection mechanism~\cite{madaan2023self}. The teacher LLM first generates candidate bundles, then compares them against ground-truth bundles to identify errors and their underlying causes. Finally, it summarizes explicit rules such as \emph{``each bundle should contain at least two items,'' ``products must align with user intent,''} and \emph{``avoid combining unrelated categories.''} These distilled rules capture general principles of bundling that can regularize the generation process and prevent common mistakes.  

    \item \textbf{Fine-grained Knowledge.} In contrast, such knowledge emphasizes session-specific reasoning. We prompt the teacher LLM to generate chain-of-thought (CoT)~\cite{wang2023self} explanations for why certain items should be bundled together, conditioned on product titles, categories, and user intents within a specific session. It may output explanations like \emph{``Customers who purchase a GPU, CPU, and motherboard are typically assembling a high-performance computer system''}, which provides fine-grained, context-aware rationales to help the student LLM align bundles with the immediate user intent.  
\end{itemize}

By combining these two types of knowledge, we cover both the generalizable rules that ensure logical and consistent bundle generation, and the session-specific reasoning that tailors predictions to specific sessions. This complementary design establishes a solid foundation for training knowledge-specific LoRA experts in the next stage.

\subsection{Knowledge-specific LoRA Experts}
As discussed in Section~\ref{sec:intro}, directly merging all types of distilled knowledge to fine-tune LLMs does not consistently yield better performance than using a single knowledge type (see Figure~\ref{fig:knowledge_comb}). This could be attributed to knowledge conflicts, where naively combining all knowledge into a single context may overwhelm the model and lead to confusion. To this end, we adopt a modular design and introduce \textit{knowledge-specific LoRA experts}, where each expert is dedicated to internalizing a distinct type of knowledge. This design separates reasoning skills into specialized LoRA modules, which can later be integrated through our dynamic fusion strategy.  

In particular, to seamlessly accommodate the distilled knowledge, our RouteDK comprises different experts.
\begin{itemize}
    \item \textbf{Base Expert} ($e_{\text{B}}$) models the intrinsic, context-independent relationships between items. It functions as the system’s core reasoning engine and as a stable backbone under noisy, missing, or irrelevant distilled knowledge, ensuring reliable bundle predictions across diverse scenarios.

    \item \textbf{High-level Expert} ($e_{\text{H}}$) functions as a strategic guide, injecting generalizable and cross-session rules into bundle generation. Its role is to ensure that generated bundles are logically consistent and aligned with broad principles, thereby regularizing the process and preventing overfitting to session-specific noise.  

    \item \textbf{Fine-grained Expert} ($e_{\text{F}}$) specializes in personalization which focuses on session-specific reasoning and subtle item interactions, refining bundles to align precisely with user intents. Unlike the High-level Expert, which emphasizes global regularities, the Fine-grained Expert excels at capturing specific user behaviors and contextual patterns within a session, enabling the model to produce highly adaptive and user-tailored bundles.
\end{itemize}

For each expert, we assign a dedicated LoRA module to the frozen backbone LLM and optimize it using a maximum likelihood estimation (MLE) objective. All experts are optimized to maximize the likelihood of generating the ground-truth textual sequence that serializes the complete set of bundles, where each bundle is represented by its containing item IDs. The optimization objective for each expert $e_i \in \mathcal{E}, \text{where}\ \mathcal{E}=\{e_{\text{B}}, e_{\text{H}}, e_{\text{F}}\}\ \text{means the expert set}$, isgiven by:
\begin{equation}\label{eq:expert}
\max_{\boldsymbol{\phi}_{e_i}} \sum\nolimits_{s \in \mathcal{S}} \sum\nolimits_{\tau=1}^{|y^s|} 
\log p(y^s_{\tau} \mid y^s_{<\tau}, x_{e_i}^s; \boldsymbol{\theta}, \boldsymbol{\phi}_{e_i}),
\end{equation}
where:
\begin{itemize}
    \item $\boldsymbol{\theta}$ represents the parameters of the frozen backbone LLM.
    \item $\boldsymbol{\phi}_{e_i}$ represents the trainable parameters of the $i$-th LoRA expert module.
    \item $y^s$ denotes the ground-truth textual sequence that serializes the complete set of bundles for session $s$ (e.g., `\{bundle1: [item1 ID, item3 ID, item4 ID], bundle2: [item2 ID, item5 ID]\}').\footnote{When a session contains multiple bundles, we construct the ground-truth textual sequence using different bundle permutations, ensuring that the student LLM learns order-invariant patterns across bundles.}
    \item $x_{e_i}^s$ is the expert-specific input for session $s$. The key distinction lies in this input:
    \begin{itemize}
        \item For the Base Expert, the input ${x}_{e_\text{B}}^s=x^s$ contains only the raw session information, such as the metadata (e.g., title, category) of the items involved. 
        \item For the High-level and Fine-grained Experts, the input ${x}_{e_\text{H}}^s\ \text{and}\ {x}_{e_\text{F}}^s$ are augmented versions of $x^s$, where the corresponding distilled knowledge (high-level or fine-grained knowledge) is injected as context.
    \end{itemize}
\end{itemize}
The model is trained in an autoregressive manner, where each token ${y}_\tau^s$ is generated sequentially conditioned on the ground-truth prefix ${y}_{<\tau}^s$ and the expert-specific input ${x}_{e_i}^s$.

\subsection{Dynamic Fusion Strategy}

With the knowledge-specific LoRA experts trained, the next challenge is to effectively integrate their outputs. A simple averaging of expert contributions is inadequate, as it cannot resolve the knowledge conflicts that arise from different types of distilled knowledge. As such, we propose an input-aware \textit{Dynamic Fusion} strategy.  
The motivation is that the optimal weights of experts are not fixed across sessions. Some sessions may benefit more from the abstract, strategic guidance of the High-level Expert, while others require the fine-grained, contextual reasoning of the Fine-grained Expert. This variability calls for a mechanism that can adaptively balance the contributions of experts according to the input.  

To realize this, we design a lightweight router network, parameterized by $\boldsymbol{\psi}$, that dynamically evaluates expert contributions at each layer. The process begins with the same raw, knowledge-free input $x^s$ as the Base Expert. The text is tokenized into a sequence of length $T$ and then passed through the frozen embedding layer of the backbone LLM, yielding the initial representation $\mathbf{H}_{\mathrm{dyn}}^0= \mathbf{X}^s \in \mathbb{R}^{T \times d}$, where $d$ is the embedding dimension. From this starting point, the router operates sequentially at each modified layer $l$, For a given layer $l$, it first receives the output of the previous layer, $\mathbf{H}_{\mathrm{dyn}}^{l-1}$, as its input. Since the router requires a fixed-length representation, a context vector $\mathbf{z}^l$ specific to the current layer's state is derived via the average pooling:
\begin{equation}
    \mathbf{z}^l=\text{AvgPool}(\mathbf{H}^{l-1}_{\text{dyn}}),\ \mathbf{z}^l\in\mathbb{R}^d.
\end{equation}
This layer-specific vector $\mathbf{z}^l$ is then processed by the router's linear transformation followed by a Softmax function to produce the weights of LoRA experts exclusively at layer $l$:
\begin{equation}
\boldsymbol{\alpha}^l(\mathbf{z}^l)=\mathrm{Softmax}\ \!\bigl(\boldsymbol{W_{\psi}}^l\mathbf{z}^l + \boldsymbol{b_{\psi}}^l\bigr).
\end{equation}
The dynamically fused output at layer $l$, $\mathbf{H}_{\mathrm{dyn}}^{l}$, is then computed by combining the backbone LLM's output with the weighted contributions from the LoRA experts:
\begin{equation}
\mathbf{H}^l_{\mathrm{dyn}}=f_{\boldsymbol{\theta}}^l(\mathbf{H}^{l-1}_{\mathrm{dyn}})+\sum\nolimits_{e_i\in \mathcal{E}} \boldsymbol{\alpha}_{e_i}^l(\mathbf{z}^l)\,\Delta_{e_i}^l(\mathbf{H}^{l-1}_{\mathrm{dyn}}),
\end{equation}
where $f_{\boldsymbol{\theta}}^l$ represents the original frozen layer and $\Delta_{e_i}^l$ is the update from LoRA expert $e_i$ at layer $l$. This layer-wise routing strategy enables granular control by dynamically determining which expert's knowledge is most relevant at each layer, adapting the fusion approach to the specific characteristics of different layers during inference.

\smallskip\noindent\textbf{Fusion Training.} 
We perform a separate training where the backbone LLM ($\boldsymbol{\theta}$) and pre-trained LoRA experts ($\boldsymbol{\phi}_{e_i}$) are frozen, and only the router parameters $\boldsymbol{\psi}$ are optimized for \textit{Dynamic Fusion}. This design prevents catastrophic forgetting of the specialized knowledge learned by individual LoRA experts, ensuring that the router focuses on learning an effective fusion strategy.
The training objective is to maximize the likelihood of the ground-truth textual sequence ${y}^s$ that serializes the complete set of bundles in the session, formulated as:
\begin{equation}
\max_{\boldsymbol{\psi}} \sum\nolimits_{s \in \mathcal{S}} \sum\nolimits_{\tau=1}^{|{y}^s|} \log p({y}_{\tau}^s | {y}_{<\tau}^s, x^s; \boldsymbol{\theta}, \{\boldsymbol{\phi}_{e_i}\}_{e_i\in \mathcal{E}}, \boldsymbol{\psi}).
\end{equation}

\subsection{Inference-time Enhancement}

Our complementary knowledge distillation, combining knowledge-specific LoRA experts and dynamic fusion strategy, yields a strong generative model for bundle generation. However, single-pass decoding at inference may still suffer from stochastic variability and occasional suboptimal reasoning. To further stabilize predictions, we design an inference-time enhancement strategy based on \textit{Test-Time Scaling (TTS)}, which is inspired by the principle of self-consistency~\cite{wang2023self}. 

Specifically, for each session with raw textual input $x^s$, we perform $N$ parallel decoding runs, each producing a complete textual sequence that encodes the predicted bundles for the session. To encourage diversity while maintaining quality, stochastic decoding (i.e., temperature sampling) is applied. This setting allows multiple runs to occasionally produce different candidate bundle sequences, while also increasing the likelihood that high-probability solutions are reproduced across runs.
To enable reliable comparison across candidates, we normalize each predicted bundle by sorting the item IDs within it in ascending order, thereby eliminating variations due to item ordering. After normalization, we apply a \textbf{majority voting} scheme over the $N$ candidate sequences and select the one that appears most frequently. This procedure effectively averages out sampling noise and captures the model’s internal distributional consensus, leading to more stable and accurate predictions. Although the computational cost increases linearly with $N$, the resulting gains in prediction reliability generally outweigh this overhead.

\section{Experiments and Results}
\subsection{Experimental Setup}

\smallskip\noindent\textbf{Datasets.} We conduct experiments on three public bundle datasets released in the SIGIR 2022 resource paper~\cite{sun2022revisiting,sun2024revisiting}. While other public datasets such as Netease~\cite{he2020consistency}, Youshu~\cite{avny2022bruce}, and iFashion~\cite{ren2023distillation} are also commonly used in product bundling research, we choose the SIGIR datasets for two important reasons. First, they are annotated via well-designed crowd-sourcing tasks, which ensures high-quality bundle labels. Second, the datasets are constructed based on Amazon datasets~\cite{he2016ups} and include rich meta-information, such as product images, textual descriptions, and user interaction data. Most importantly, they contain session-level information, capturing all items a user interacts within a session. This makes them particularly suitable for the objectives of our study. The statistics of the datasets are summarized in Table~\ref{tab:datasets}. For each dataset, we chronologically split the session data into training, validation, and test sets with a ratio of 7:1:2. 
%

\begin{table}[t]
    \centering
    \small
    \caption{The statistics of the three bundle datasets.}
    \label{tab:datasets}
    \addtolength{\tabcolsep}{2pt}
    \vspace{-0.1in}
    \begin{tabular}{c|c|c|c}
    \toprule
    &Electronic &Clothing &Food \\\midrule
    \#Users  &888 &965 &879\\
    \#Items &3499 &4487 &3767\\
    \#Sessions &1145 &1181 &1161\\
    \#Bundles &1750 &1910 &1784\\
    \#Intents &1422 &1466 &1156\\
    \#User-Item Interactions &6165 &6326 &6395\\ 
    \#User-Bundle Interactions &1753 &1912 &1785\\
    Average Bundle Size &3.52 &3.31 &3.58 \\
    \bottomrule
    \end{tabular}
    \vspace{-0.1in}
\end{table}


\smallskip\noindent\textbf{Baselines.} 
We evaluate our proposed model against two types of methods: traditional-based and LLM-based methods. The traditional ones include: \textit{Freq}~\cite{sun2022revisiting}, which uses the Apriori algorithm to identify frequent patterns at the item category level; \textit{BBPR}~\cite{pathak2017generating}, a greedy algorithm that dynamically generates bundles based on the similarities between representations of users, items, and bundles learned through BPRMF~\cite{rendle2012bpr}; and \textit{POG}~\cite{chen2019pog}, a Transformer-based model that creates personalized outfits from multi-modal data. 
For LLM-based methods, we treat \textit{GPT-3.5-turbo} as the teacher LLM and report its zero-shot performance on the bundle generation task. We also compare with \textit{AICL}\cite{sun2024adaptive}, a retrieval-augmented method built on GPT-3.5-turbo that achieves state-of-the-art results. For the student LLM, we adopt the open-source \textit{Llama3.1-8B-Instruct}, reporting its zero-shot performance and two enhanced variants from\cite{feng2025does}: a \textit{Supervised Fine-Tuning (SFT w/o KD)} model trained on the raw dataset only, and a \textit{Supervised Fine-Tuning with Knowledge Distillation (SFT w/ KD)} model trained on the raw dataset augmented with distilled knowledge from the teacher LLM.
The \textit{SFT w/ KD} variant is the state-of-the-art under KD-based LLMs for bundle generation.


\smallskip\noindent\textbf{Evaluation Metrics.} {We adopt the same evaluation metrics as in prior work~\cite{sun2022revisiting, sun2024revisiting, sun2024adaptive, feng2025does} to evaluate the quality of generated bundles. Specifically, at the session level, we adopt \textit{Precision} and \textit{Recall} to measure the quantity of correctly predicted bundles within each session. \textit{Precision} indicates how many of the generated bundles are correct, while \textit{Recall} measures how many of the ground truth bundles are identified. Note that we treat a generated bundle as a hit bundle if it either completely matches or is a subset of a ground truth bundle. At the bundle level, we use \textit{Coverage} to evaluate the item-wise accuracy of hit bundles by calculating the average proportion of correctly predicted items within the ground truth bundles.}

\smallskip\noindent\textbf{Implement Details}.
For traditional methods, we adopt the parameter configurations reported in~\cite{sun2024adaptive}. For LLM-based methods, we follow the setup in~\cite{feng2025does}. The teacher LLM and AICL are built upon GPT-3.5-turbo, while the student LLM is based on Llama3.1-8B-Instruct. For the two variants in~\cite{feng2025does} and our LoRA experts, we use a learning rate of 2e-4, 3 training epochs, and set both rank and scaling factor to 16. All fine-tuning is performed with QLoRA implemented in the Unsloth framework (https://unsloth.ai/). For our Dynamic Fusion strategy, we grid search learning rates in \{1e-4, 1e-3, 1e-2\}. In test-time scaling, we search the number of candidates in the range of \{4, 8, 16, 32\}. During inference, the temperature is set to 0 by default to ensure deterministic decoding, while for test-time scaling it is adjusted to 0.7 to promote output diversity, following prior work~\cite{li2025s}. All experiments are conducted on 4× NVIDIA A40 GPUs (48GB), with a batch size of 2 and a gradient accumulation step of 4.
\begin{table*}[t]
\centering
\small  
\caption{Effectiveness comparison across different methods in the three domains.}
\vspace{-0.1in}
\setlength{\tabcolsep}{4.5pt}  
\begin{tabular}{l|c|ccc|ccc|ccc}
\toprule
\multirow{2}{*}{Type} & \multirow{2}{*}{Method} & \multicolumn{3}{c|}{Electronic} & \multicolumn{3}{c|}{Clothing} & \multicolumn{3}{c}{Food} \\
\cline{3-11}
& & Precision & Recall & Coverage & Precision & Recall & Coverage & Precision & Recall & Coverage \\
\midrule
\multirow{4}{*}{\begin{tabular}[c]{@{}l@{}}Traditional\\ Methods\end{tabular}} & Freq & 0.423 & 0.597 & 0.701 & 0.532 & 0.566 & 0.698 & 0.491 & 0.525 & 0.684 \\
 & BBPR & 0.260 & 0.122 & 0.433 & 0.239 & 0.211 & 0.449 & 0.210 & 0.183 & 0.416 \\
 & POG & 0.339 & 0.250 & 0.412 & 0.312 & 0.221 & 0.399 & 0.365 & 0.266 & 0.393 \\
 & BYOB & 0.340 & 0.294 & 0.361 & 0.311 & 0.273 & 0.457 & 0.304 & 0.253 & 0.427 \\
\midrule
\multirow{5}{*}{\begin{tabular}[c]{@{}l@{}}LLM-based\\Methods\end{tabular}}
& Student LLM & 0.564 & 0.582 & 0.685 & 0.582 & 0.644 & 0.655 & 0.571 & 0.591 & 0.633 \\
& $+$ SFT w/o KD~\cite{feng2025does} & 0.618 & 0.621 & 0.857 & 0.608 & 0.607 & 0.901 & 0.615 & 0.597 & 0.841 \\
& \cellcolor{gray!30}$+$ SFT w/ KD~\cite{feng2025does}& \cellcolor{gray!30}0.650 & \cellcolor{gray!30}0.682 & \cellcolor{gray!30}0.783 & \cellcolor{gray!30}0.665 & \cellcolor{gray!30}0.625 & \cellcolor{gray!30}0.832 & \cellcolor{gray!30}{0.707} & \cellcolor{gray!30}0.681 & \cellcolor{gray!30}0.834 \\
&Teacher LLM & 0.580 & 0.820 & 0.720 & 0.603 & 0.752 & 0.788 & 0.604 & 0.815 & 0.748 \\
& AICL~\cite{sun2024adaptive} & 0.769 & 0.859 & 0.741 & 0.677 & 0.788 & 0.839 & 0.698 & 0.851 & 0.755 \\
\midrule
\multirow{1}{*}{\begin{tabular}[c]{@{}l@{}}Ours\end{tabular}}
 & RouteDK & \textbf{0.734} & \textbf{0.748} & \textbf{0.808} & \textbf{0.758} & \textbf{0.745} & \textbf{0.881} & \textbf{0.761} & \textbf{0.756} & \textbf{0.837} \\
\midrule
\multicolumn{2}{l|}{\textit{Improve}} &12.92\% &9.68\% &3.19\% &13.98\% &19.20\%&5.89\% &7.64\% &11.01\% &0.35\% \\
\bottomrule
\end{tabular}\label{tab:baselines}
\vspace{-0.15in}
\end{table*}

\subsection{Performance Comparison}

Table~\ref{tab:baselines} reports the performance of all methods across all domains and metrics. The results of +SFT w/ KD~\cite{feng2025does}, which represents the state-of-the-art approach for KD-based LLMs in the bundle generation task, are highlighted in gray. The results of our RouteDK are marked in bold, and the row `\textit{Improve}' quantifies the relative improvements of RouteDK over the results in gray. Several major findings are noted.

(1) \textit{\textbf{Our proposed RouteDK consistently outperforms traditional bundle generation methods across all datasets and metrics.}} Compared with all traditional-based methods, RouteDK achieves markedly higher Precision, Recall, and Coverage.  This demonstrates that even small-scale LLMs with KD brings substantial gains over conventional approaches.

(2) \textit{\textbf{RouteDK achieves performance comparable to, and in some cases better than, large teacher LLMs.}} Despite being trained on the smaller Llama3.1-8B-instruct, RouteDK narrows the gap with the teacher LLMs (GPT-3.5-turbo and AICL). Notably, it surpasses AICL in Precision and Coverage for both the Clothing and Food domains, and consistently achieves the best Coverage across all domains. These findings highlight that lightweight student LLMs, when properly equipped with distilled knowledge from the teacher LLM can rival or even outperform large commercial LLMs while retaining efficiency.

(3) \textit{\textbf{Among student LLM-based methods, RouteDK achieves the best overall results.}} Compared with +SFT w/o KD and +SFT w/ KD methods, RouteDK delivers significant improvements, particularly in Precision and Recall. For example, on the Electronic dataset, RouteDK improves Precision and Recall by 12.92\% and 9.68\% over +SFT w/ KD, while on Clothing the gains reach 13.98\% and 19.20\%. Notably, +SFT w/o KD achieves relatively higher Coverage than RouteDK. This stems from the fact that Coverage only measures item coverage within successfully predicted bundles. +SFT w/o KD tends to predict only simple bundles while failing to capture more complex ones, resulting in lower Precision and Recall but higher Coverage.
These results showcase the effectiveness of decoupling different types of knowledge into specialized experts and combining multiple LoRA experts through the dynamic fusion strategy to better resolve knowledge conflicts and enhance predictive performance on bundle generation task.

\subsection{Ablation Study}

\smallskip\noindent\textbf{Impact of Knowledge-specific Experts.} To validate the necessity of the knowledge-specific expert module, we conduct an ablation study that examines whether separating knowledge into specialized experts is indeed necessary. A simple alternative is to collapse all types of knowledge with the raw data into a single LoRA module ($K$=1), which forces the model to learn from mixed signals without differentiation. Another compromise is to split the raw data from external knowledge but still merge the high-level and fine-grained knowledge into one LoRA module ($K$=2). Our proposed approach assigns three independent experts ($K$=3), each dedicated to raw data, high-level knowledge, or fine-grained knowledge. 

The results, shown in Figure~\ref{fig:n_experts}, demonstrate a clear trend: both Precision and Recall consistently improve as the number of specialized experts increases. More importantly, the largest gain comes from moving from $K$=2 to $K$=3, indicating that high-level and fine-grained knowledge cannot be effectively captured within a shared module. This provides strong evidence that expert specialization is essential for mitigating knowledge conflicts and allowing the model to capture the complementary strengths of different knowledge types.

\pgfplotsset{colormap={blues}{
    rgb255(0.00)=(255,255,204)
    rgb255(0.25)=(161,218,180)
    rgb255(0.50)=(65,182,196)
    rgb255(0.75)=(44,127,184)
    rgb255(1.00)=(37,52,148)
}}

\begin{figure}[tbp]
    \centering
    \begin{subfigure}[b]{0.45\columnwidth}
        \centering
        \begin{tikzpicture}
            \begin{axis}[
                width=4.5cm,
                height=4.5cm,
                colorbar,
                colormap name=blues,     
                view={0}{90},
                xmin=-0.5, xmax=2.5,
                ymin=-0.5, ymax=2.5,
                title={Precision},
                xtick={0,1,2},
                xticklabels={Elec,Cloth,Food},
                ytick={0,1,2},
                yticklabels={$K$=3,$K$=2,$K$=1},
                ticklabel style={font=\footnotesize},
                axis equal,
                colorbar style={
                    width=5pt,
                    title style={yshift=5pt}
                }
            ]
            \addplot3[
                matrix plot,
                mesh/rows=3,
                mesh/cols=3,
                point meta=explicit
            ] table[meta=z] {
                x y z
                0 0 0.737
                1 0 0.758
                2 0 0.760
                0 1 0.682 
                1 1 0.732 
                2 1 0.723
                0 2 0.610
                1 2 0.663
                2 2 0.665
            };
            \end{axis}
        \end{tikzpicture}
    \end{subfigure}%
    \hfill
    \begin{subfigure}[b]{0.45\columnwidth}
        \centering
        \begin{tikzpicture}
            \begin{axis}[
                width=4.5cm,
                height=4.5cm,
                colorbar,
                colormap name=blues,  
                view={0}{90},
                xmin=-0.5, xmax=2.5,
                ymin=-0.5, ymax=2.5,
                title={Recall},
                xtick={0,1,2},
                xticklabels={Elec,Cloth,Food},
                xticklabel style={font=\footnotesize},
                ytick={0,1,2},
                yticklabels={,,},
                ticklabel style={font=\footnotesize},
                axis equal,
                colorbar style={
                    width=5pt,
                    title style={yshift=5pt}
                }
            ]
            \addplot3[
                matrix plot,
                mesh/rows=3,
                mesh/cols=3,
                point meta=explicit
            ] table[meta=z] {
                x y z
                0 0 0.748
                1 0 0.745
                2 0 0.755
                0 1 0.699
                1 1 0.717
                2 1 0.713
                0 2 0.648
                1 2 0.674
                2 2 0.684
            };
            \end{axis}
        \end{tikzpicture}
    \end{subfigure}
    \caption{The impact of the number of LoRA experts ($K$) in the three domains.}
    \label{fig:n_experts}
    \vspace{-0.2in}
\end{figure}
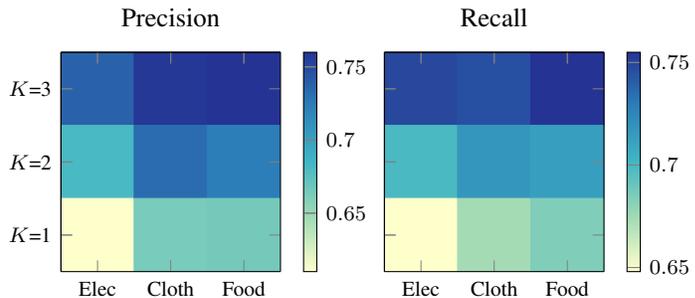

\smallskip\noindent\textbf{Impact of the Fusion Strategies.} 
We further examine the effect of different strategies for fusing the outputs of multiple experts. Specifically, we compare four approaches: (i) \textit{Average Pooling}, which directly averages the outputs of all experts at layer $l$, formalized as $\mathbf{H}_\mathrm{ave}^l= f_{\boldsymbol{\theta}}^l(\mathbf{H}^{l-1}_{\mathrm{ave}})+\frac{1}{|\mathcal{E}|}\sum_{e_i\in \mathcal{E}}\Delta_{e_i}^l(\mathbf{H}_\mathrm{ave}^{l-1})$; (ii) \textit{TIES}~\cite{yadav2023ties}, a sparsity-guided, sign-voting method for merging LoRA adapters that preserves the most impactful updates, resolves conflicting weights, and improves efficiency; (iii) \textit{Static}, an input-agnostic strategy that learns layer-wise coefficients ${\gamma}_{e_i}^l$ for each expert $e_i$ at layer $l$. This fusion strategy is expressed as $\mathbf{H}_\mathrm{sta}^l=f_{\boldsymbol{\theta}}^l(\mathbf{H}^{l-1}_{\mathrm{sta}})+\sum_{e_i\in \mathcal{E}}\gamma_{e_i}^l\,\Delta_{e_i}^l(\mathbf{H}_\mathrm{sta}^{l-1})$ and these coefficients remain fixed during inference; (iv) \textit{Dynamic}, our proposed input-aware strategy that generates adaptive weights conditioned on the input.  

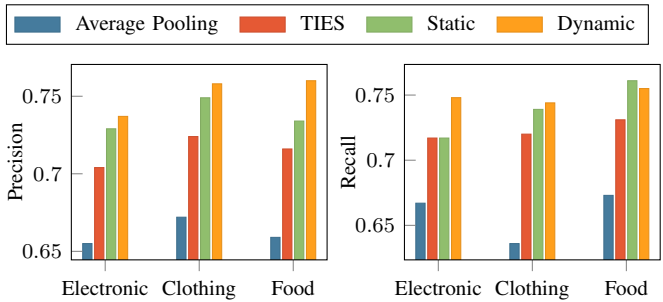
\begin{figure}[tbp]
    \centering
    \vspace{-0.00in}
    \pgfplotsset{
        commonaxis/.style={
            width=3.4cm,          
            height=2.6cm,
            xtick={1,2,3},
            axis on top,
            extra y ticks={0},
            extra y tick style={
                grid=major,
                grid style={densely dashed, line width=0.5pt, black!50},
            },
            xticklabels={Electronic, Clothing, Food},
            xticklabel style={font=\footnotesize},
            enlarge x limits={abs=0.45cm},
            scaled ticks=false,
            xtick pos=left,
            ytick pos=left,
            yticklabel style={
            /pgf/number format/fixed,
            /pgf/number format/precision=3,  
            font=\footnotesize
            },
            grid=none,
            bar width=3.5pt,
            ybar=1pt,
            scale only axis,
            ylabel style={font=\footnotesize},
            ylabel shift=-7pt
        },
    }
    
    \begin{tikzpicture}
        \begin{axis}[
            hide axis,
            scale only axis,
            height=0pt,
            width=0pt,
            xmin=0, xmax=1,
            ymin=0, ymax=1,
            legend style={
                font=\footnotesize,
                legend columns=6,
                column sep=0.8ex,
                /tikz/every even column/.append style={column sep=0.6em}
            }
        ]
        \addlegendimage{area legend, fill=color1, draw=color1!90!black}
        \addlegendimage{area legend, fill=color2, draw=color2!90!black}
        \addlegendimage{area legend, fill=color3, draw=color3!90!black}
        \addlegendimage{area legend, fill=color4, draw=color4!90!black}
        
        \legend{Average Pooling, TIES, Static, Dynamic}
        \end{axis}
    \end{tikzpicture}
    \vspace{0.1in}

    \begin{subfigure}[t]{0.5\columnwidth}
        \begin{tikzpicture}
            \begin{axis}[commonaxis, ylabel={Precision}]
                \addplot[fill=color1, draw=color1!90!black] coordinates {(1, 0.655) (2, 0.672) (3, 0.659)};
                \addplot[fill=color2, draw=color2!90!black] coordinates {(1, 0.704) (2, 0.724) (3, 0.716)};
                \addplot[fill=color3, draw=color3!90!black] coordinates {(1, 0.729) (2, 0.749) (3, 0.734)};
                \addplot[fill=color4, draw=color4!90!black] coordinates {(1, 0.737) (2, 0.758) (3, 0.760)};
            \end{axis}
        \end{tikzpicture}
    \end{subfigure}%
    \hfill
    \begin{subfigure}[t]{0.5\columnwidth}
        \begin{tikzpicture}
            \begin{axis}[commonaxis, ylabel={Recall}]
                \addplot[fill=color1, draw=color1!90!black] coordinates {(1, 0.667) (2, 0.636) (3, 0.673)};
                \addplot[fill=color2, draw=color2!90!black] coordinates {(1, 0.717) (2, 0.720) (3, 0.731)};
                \addplot[fill=color3, draw=color3!90!black] coordinates {(1, 0.717) (2, 0.739) (3, 0.761)};
                \addplot[fill=color4, draw=color4!90!black] coordinates {(1, 0.748) (2, 0.744) (3, 0.755)};
            \end{axis}
        \end{tikzpicture}
    \end{subfigure}
    
    \caption{The impact of different fusion strategies across the three domains.}
    \vspace{-0.1in}
    \label{fig:merging_strategy}
\end{figure}

As illustrated in Figure~\ref{fig:merging_strategy}, the Dynamic strategy generally achieves the best Precision and Recall across domains, highlighting the advantage of context-aware fusion. A slight exception occurs in the Food domain, where Static surpasses Dynamic in Recall, suggesting that globally learned coefficients can sometimes provide more stable guidance when domain signals are highly consistent. Nonetheless, both trainable strategies (Static and Dynamic) consistently outperform non-trainable ones (Average Pooling and TIES). Meanwhile, TIES improves over Average Pooling by filtering out conflicting updates and those with minimal contribution at the model parameter level, though its gains remain moderate compared to trainable strategies. Overall, the results indicate a clear trend: trainable, data-driven fusion (Static and Dynamic) is crucial for enabling multiple experts to collaborate effectively, with Dynamic offering the strongest performance in most cases.  

\begin{table*}[t]
\centering
\small  
\caption{The effectiveness of inference-time enhancement strategy in the three domains.}
\vspace{-0.1in}
\addtolength{\tabcolsep}{2pt}
\begin{tabular}{c|ccc|ccc|ccc}
\toprule
\multirow{2}{*}{Method} & \multicolumn{3}{c|}{Electronic} & \multicolumn{3}{c|}{Clothing} & \multicolumn{3}{c}{Food} \\
\cline{2-10}
& Precision & Recall & Coverage & Precision & Recall & Coverage & Precision & Recall & Coverage \\
\midrule  
RouteDK\_s & 0.729 & 0.717 & 0.789 & 0.749 & 0.739 & 0.867 & 0.734 & 0.761 & 0.811 \\
$-$ TTS & 0.724 & 0.713 & 0.785 & 0.733 & 0.735 & 0.871 & 0.731 & 0.721 & 0.815 \\
$+$ RAC & 0.658 & 0.638 & 0.829 & 0.662 & 0.625 & 0.873 & 0.631 & 0.609 & 0.821 \\
\midrule
RouteDK\_d & 0.734 & 0.748 & 0.808 & 0.758 & 0.745 & 0.881 & 0.761 & 0.756 & 0.837 \\
$-$ TTS & 0.728 & 0.726 & 0.778 & 0.739 & 0.738 & 0.869 & 0.752 & 0.742 & 0.807 \\
$+$ RAC & 0.624 & 0.622 & 0.829 & 0.644 & 0.605 & 0.861 & 0.641 & 0.623 & 0.803 \\

\bottomrule
\end{tabular}\label{tab:inference}
\vspace{-0.15in}
\end{table*}

\smallskip\noindent\textbf{Impact of Inference-time Enhancement.} 
We further analyze the role of inference-time enhancement strategy in improving model performance. \textbf{First}, we investigate the effectiveness of our proposed \textit{TTS with majority voting}. Table~\ref{tab:inference} reports the impact of inference-time enhancement strategy on both the static fusion (RouteDK\_s) and the dynamic fusion (RouteDK\_d) across three domains. We observe that removing TTS leads to consistent drops in accuracy and stability across all datasets, confirming its necessity for robust predictions. We also compare with a \textit{Retrieval-Augmented Context (RAC)} strategy, which retrieves the most similar session based on SentenceTransformer embeddings and incorporates its distilled knowledge into the test input. Surprisingly, RAC causes performance degradation, especially under RouteDK\_d. This suggests that introducing external fine-grained knowledge from other sessions may introduce conflicts with the target session, imposing overly rigid constraints that misguide the fine-grained expert.

\begin{figure}[tbp]
    \centering
    \vspace{0.05in}
    \pgfplotsset{
        commonaxis/.style={
            width=3cm,          
            height=2.2cm,
            xtick={1,2,3,4,5},
            axis on top,
            extra y ticks={0},
            xticklabels={1,4,8,16,32},
            xticklabel style={font=\footnotesize},
            enlarge x limits={abs=0.45cm},
            scaled ticks=false,
            xtick pos=left,
            ytick pos=left,
            yticklabel style={
            /pgf/number format/fixed,
            /pgf/number format/precision=3,  
            font=\footnotesize
            },
            scaled ticks=false,
            grid=none,
            bar width=3.5pt,
            ybar=1pt,
            scale only axis,
            label style={font=\footnotesize},
            ylabel shift=-3pt
        },
    }
    
    \begin{tikzpicture}
        \begin{axis}[
            hide axis,
            scale only axis,
            height=0pt,
            width=0pt,
            xmin=0, xmax=1,
            ymin=0, ymax=1,
            legend style={
                font=\footnotesize,
                legend columns=6,
                column sep=0.8ex,
                /tikz/every even column/.append style={column sep=1em}
            }
        ]
        \addlegendimage{line legend, mark=*, fill=color1, draw=color1!90!black}
        \addlegendimage{line legend, mark=*, fill=color3, draw=color3!90!black}
        \addlegendimage{line legend, mark=*, fill=color4, draw=color4!90!black}
        
        \legend{Electronic, Clothing, Food}
        \end{axis}
    \end{tikzpicture}
    \vspace{0.05in}

    \begin{subfigure}[t]{0.5\columnwidth}
        \begin{tikzpicture}
            \begin{axis}[commonaxis, ylabel={Precision}, xlabel={$N$}]
                \addplot[color=color1, mark=*, solid, line width=1pt, smooth] coordinates {(1, 0.728) (2, 0.721) (3, 0.734) (4, 0.725) (5, 0.717)};
                \addplot[color=color3, mark=square, solid, line width=1pt, smooth] coordinates {(1, 0.739) (2, 0.729) (3, 0.758) (4, 0.739) (5, 0.734)};
                \addplot[color=color4, mark=diamond, solid, line width=1pt, smooth] coordinates {(1, 0.752) (2, 0.740) (3, 0.761) (4, 0.755) (5, 0.753)};

            \end{axis}
        \end{tikzpicture}
    \end{subfigure}%
    \hfill
    \begin{subfigure}[t]{0.5\columnwidth}
        \begin{tikzpicture}
            \begin{axis}[commonaxis, ylabel={Recall}, xlabel={$N$}]
                \addplot[color=color1, mark=*, solid, line width=1pt, smooth] coordinates {(1, 0.726) (2, 0.731) (3, 0.748) (4, 0.734) (5, 0.729)};
                \addplot[color=color3, mark=square, solid, line width=1pt, smooth] coordinates {(1, 0.738) (2, 0.750) (3, 0.745) (4, 0.740) (5, 0.731)};
                \addplot[color=color4, mark=diamond, solid, line width=1pt, smooth] coordinates {(1, 0.742) (2, 0.750) (3, 0.756) (4, 0.746) (5, 0.750)};

            \end{axis}
        \end{tikzpicture}
    \end{subfigure}
    
    \caption{The impact of different sampling number $N$ across the three domains.}
    \vspace{-0.1in}
    \label{fig:TTS_N}
\end{figure}
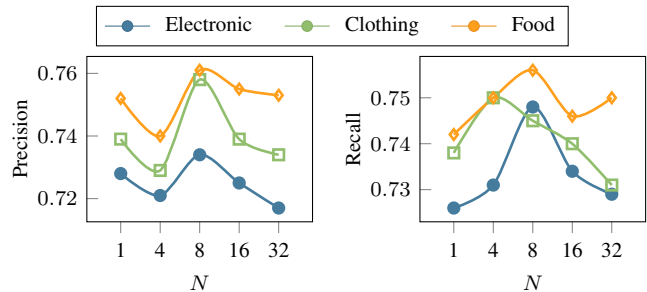

\textbf{Second}, we study the impact of the sampling number $N$ in test-time scaling. Figure~\ref{fig:TTS_N} reports Precision and Recall across three datasets with varying $N$. We find that performance steadily improves as $N$ increases from 1 to 8, but then declines, with clear degradation at $N$=32. This trend indicates that excessive sampling introduces noisy candidates, which reduces the effectiveness of majority voting. Based on these observations, we select $N$=8 as the optimal configuration, as it strikes a practical balance between performance gains and computational efficiency.

\subsection{Case Study}

\begin{figure}
    \centering
    \includegraphics[width=0.9\columnwidth]{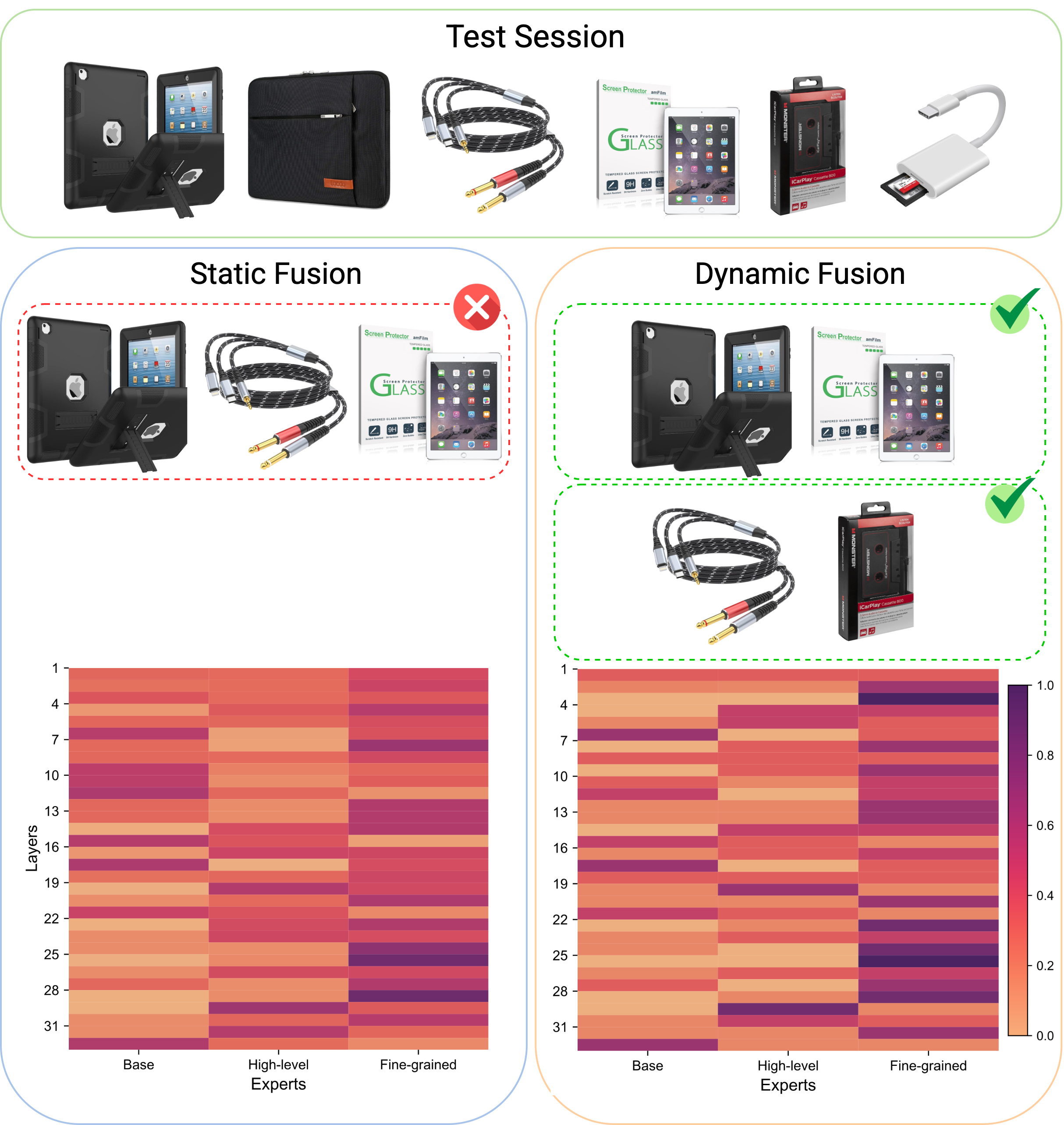}
    \caption{The generated bundles by different fusion strategies on Electronic.}
    \label{fig: case_study}
\end{figure}

To qualitatively investigate the performance of our RouteDK framework, we present a case study in Figure~\ref{fig: case_study} that visualizes expert weight distributions across model layers during inference. For a complex session from the Electronics domain containing two distinct user intents (iPad protection and car audio), our Dynamic Fusion strategy successfully identifies and generates two coherent bundles, whereas Static Fusion fails by incorrectly merging unrelated items. The expert-weight heatmap across layers illustrates the underlying reason:
our dynamic router consistently assigns a dominant weight to the Fine-grained Expert across nearly all layers, recognizing that decoding a specific session requires a detailed understanding of nuanced, context-dependent item relationships. The High-level Expert serves a supporting role, likely providing semantic validation, while the Base Expert contributes minimally. In contrast, Static Fusion produces more balanced weights across the three experts, which results in insufficient fine-grained understanding. For instance, it incorrectly groups an item that should belong to car audio into iPad-related accessories simply due to the shared keyword `Apple Lightning Interface'. This case illustrates that our dynamic, layer-level routing mechanism can adapt its reasoning strategy to the complexity of the input, yielding significantly more precise and logically consistent bundle generation than static fusion approaches.

\section{Conclusions and Future Works}
In this work, we tackle the knowledge conflict problem in KD for bundle generation, where simply merging different types of knowledge can degrade performance. To overcome this, we propose RouteDK, a framework that explicitly separates knowledge acquisition and utilization. RouteDK distills two complementary types of knowledge from the teacher LLM, namely high-level knowledge capturing general rules and fine-grained knowledge capturing session specific reasoning. These distilled knowledge are then accommodated by specialized LoRA experts: Base, High-level, and Fine-grained. To effectively integrate their outputs, we design a dynamic fusion module equipped with a lightweight router that adaptively balances expert contributions based on the input to mitigate the potential knowledge conflict issue. Additionally, we adopt a test-time scaling strategy to further enhance inference robustness through self-consistency.
Extensive experiments on three public datasets validate the effectiveness of RouteDK. The results show that our framework achieves substantial improvements over state-of-the-art KD baselines, and in several cases even rivals the performance of the teacher LLM. Importantly, these gains are achieved while preserving the efficiency advantages of lightweight LLMs, demonstrating the practical value of our approach for bundle generation.

\smallskip\noindent\textbf{Limitations and Future Work.}
Despite the promising results, our work still has several limitations. \textit{\textbf{First}}, our framework only leverages textual knowledge, which underutilizes the potential of multimodal signals. Future work will explore incorporating images and item relationships to enrich the knowledge space and enhance task understanding. \textbf{\textit{Second}}, our test time scaling strategy remains simplistic, relying on majority voting that may miss optimal solutions. A promising direction is to guide models to generate explicit reasoning paths and use reward models to evaluate them, enabling more reliable answer selection. \textit{\textbf{Third}},  the current gains on efficiency are mainly achieved through parameter reduction, without structural optimization. Future work will investigate expert activation mechanisms, such as partially training expert parameters or selectively activating subsets of experts during inference, to further improve training and inference efficiency.


\section*{Acknowledgments}
This paper is supported by Open Foundation of Key Laboratory of Interdisciplinary Research of Computation and Economics (Shanghai University of Finance and Economics), Ministry of Education, China. It is partially supported by the Ministry of Education, Singapore, under its MOE AcRF Tier 1, SUTD Kickstarter Initiative (SKI 2021\_06\_12). We also greatly acknowledge the support of the National Natural Science Foundation of China (Grant No. 72371148 and 72192832), the Shanghai Rising-Star Program (Grant No. 23QA1403100), and the Program for Innovative Research Team of Shanghai University of Finance and Economics. This paper has been refined by Claude and ChatGPT to polish the language and improve the clarity of the manuscript.



 




\vfill

\end{document}